# Constrained Multi-objective Bayesian Optimization through Optimistic Constraints Estimation

**Diantong Li**
CUHK, Shenzhen

**Fengxue Zhang**
University of Chicago

**Chong Liu**
University at Albany, SUNY

**Yuxin Chen**
University of Chicago

## Abstract

Multi-objective Bayesian optimization has been widely adopted in scientific experiment design, including drug discovery and hyper-parameter optimization. In practice, regulatory or safety concerns often impose additional thresholds on certain attributes of the experimental outcomes. Previous work has primarily focused on constrained single-objective optimization tasks or active search under constraints. The existing constrained multi-objective algorithms address the issue with heuristics and approximations, posing challenges to the analysis of the sample efficiency. We propose a novel constrained multi-objective Bayesian optimization algorithm (**COMBOO**) that balances active learning of the level-set defined on multiple unknowns with multi-objective optimization within the feasible region. We provide both theoretical analysis and empirical evidence, demonstrating the efficacy of our approach on various synthetic benchmarks and real-world applications.

## 1 INTRODUCTION

Multi-objective Bayesian optimization (MBO) plays a critical role in various scientific fields, particularly where experimental efficiency and precision are paramount. In drug discovery, for example, researchers must navigate a vast experimental space, balancing multiple objectives like maximizing therapeutic efficacy while minimizing toxicity and adverse effects (Fromer and Coley, 2023). The challenge lies not only in identifying promising drug candidates but also in meeting stringent regulatory and safety requirements, which impose additional constraints on the optimization process (Mellinghoff and Cloughesy, 2022). A failure to meet these regulatory and safety constraints can lead to significant delays in clinical trials or even abandonment of potential drug candidates. Similarly, in hyper-parameter optimization for machine learning models, there is a need to balance model accuracy, recall, and robustness to distribution shift (Gardner et al., 2019). Here, efficient exploration of hyper-parameter space must also respect practical constraints, such as avoiding configurations that lead to excessively high training times or resource overuse (Karl et al., 2023). These practical considerations motivate the constrained multi-objective Bayesian optimization (Fernández-Sánchez et al., 2023; Hernández-Lobato et al., 2016) beyond MBO.

Research in Bayesian Optimization (BO) has mainly focused on unconstrained problems, with Constrained Bayesian Optimization (CBO) evolving from early work by Schonlau et al. (1998). Subsequent efforts introduced posterior sampling (Eriksson and Poloczek, 2021) and information-based methods (Hernández-Lobato et al., 2014; Wang and Jegelka, 2017) to improve scalability and feasibility analysis (Hernández-Lobato et al., 2015; Perrone et al., 2019; Takeno et al., 2022). The augmented Lagrangian framework helped convert constrained tasks into unconstrained ones (Gramacy et al., 2016; Picheny et al., 2016), though often without guarantees. More recent work (Zhou and Ji, 2022; Lu and Paulson, 2022; Xu et al., 2023; Guo et al., 2023) seeks theoretical convergence for relaxed CBO objectives. In constrained multi-objective Bayesian Optimization, generalized information-criteria-based algorithms (Hernández-Lobato et al., 2016; Fernández-Sánchez et al., 2023) allow a tradeoff between constraint learning and objective optimization but rely on necessary approximations for traceability and lack guarantees, similar to their single-objective counterparts.

Our **contributions** are summarized as follows.

Constrained Multi-objective Bayesian Optimization1. We propose a sample-efficient constrained multi-objective Bayesian optimization (CMOBO) algorithm that balances learning of level sets on multiple unknown objectives and constraints with multi-objective optimization within feasible regions. The insight is that we constrain the search space to potentially feasible areas, while the random scalarization allows an efficient and theoretically justified acquisition within the region.

2. We provide theoretical analysis on the sample efficiency and capability of declaring infeasibility. To the best of our knowledge, this is the first work in constrained multi-objective optimization that offers a guarantee similar to that of constrained single-objective algorithms.

3. We offer empirical evidence on both synthetic benchmarks and real-world applications, demonstrating the effectiveness and efficiency of the proposed method against existing baselines.

## 2 RELATED WORK

**Constrained Bayesian optimization.** Constrained Bayesian optimization primarily focuses on extending unconstrained problems to optimize a single objective under constraints. Schonlau et al. (1998) pioneered this field by extending Expected Improvement (EI) to handle black-box constraints, with subsequent refinements by Gelbart et al. (2014); Gardner et al. (2014); Feliot et al. (2017); Letham et al. (2019); Wang et al. (2024). These approaches typically define the acquisition function as the product of the expected improvement and the feasibility probability. Posterior sampling methods, such as Thompson sampling in SCBO (Eriksson and Poloczek, 2021), and information-based methods (Hernández-Lobato et al., 2014; Wang and Jegelka, 2017; Hernández-Lobato et al., 2015; Perrone et al., 2019; Takeno et al., 2022) have been adapted to constrained settings. The augmented Lagrangian framework (Gramacy et al., 2016; Picheny et al., 2016; Ariafar et al., 2019) transformed constrained BO into unconstrained tasks, though often lacking guarantees on feasible region identification and regret bounds. Recent works (Zhou and Ji, 2022; Lu and Paulson, 2022; Xu et al., 2023; Guo et al., 2023; Zhang et al., 2023b) have explored relaxed CBO objectives to enable theoretical convergence analysis, assuming queries outside the feasible region incur rewards.

**Constrained active learning.** Active learning for constraint estimation (AL-LSE) (Gotovos et al., 2013) leverages Gaussian Processes (GPs) for theoretical guarantees but struggles with handling multiple unknown functions simultaneously. Although approaches like Malkomes et al. (2021); Komiyama et al. (2022) propose acquisition functions prioritizing diversity in active search, they do not effectively balance learning constraints with optimizing the objective.

**Multi-objective optimization.** In multi-objective optimization, the learner needs to optimize multiple objectives simultaneously, presenting challenges in defining valid evaluation metrics. To address this, hypervolume, defined as the volume between the Pareto frontier and any reference point, is used as a single objective. Key developments in hypervolume-based methods include Yang et al. (2019)'s pioneering work on exact Expected HyperVolume Improvement (EHVI), Daulton et al. (2020, 2021)'s extension to parallel settings with qEHVI and qNEHVI algorithms, Ament et al. (2023)'s logEI to address vanishing acquisition values, and Daulton et al. (2022)'s MORBO for high-dimensional optimization. Konakovic Lukovic et al. (2020) proposed the diversified batch query for improving hypervolume. Golovin and Zhang (2020) provided theoretical results on hypervolume regrets. Concurrently, active learning approaches have focused on directly learning the Pareto frontier, with Zuluaga et al. (2016); Belakaria et al. (2020b) proposing uncertainty reduction methods, Suzuki et al. (2020) introducing the Pareto-Frontier Entropy Search (PFES), and Park et al. (2024) developing BOtied based on a cumulative distribution function indicator.

**Constrained multi-objective optimization.** There is limited research directly addressing constrained multi-objective optimization problems. Without Bayesian optimization, Afshari et al. (2019) studied non-Bayesian algorithms for known functions. In the context of BO, the widely available Botorch (Balandat et al., 2020) implementations of Daulton et al. (2020) and Daulton et al. (2021), though only briefly mentioning constrained scenarios, weight acquisitions with feasibility probability, similar to Gelbart et al. (2014)'s approach for constrained Expected Improvement. These methods rely on objective scalarization and constrained expected improvements. Hernández-Lobato et al. (2016) pioneered the extension of Predictive Entropy Search (PES) to constrained multi-objective optimization, explicitly balancing constraint learning and Pareto frontier discovery. To address PES's computational complexity, Fernández-Sánchez et al. (2023) advanced this concept by incorporating Max-value Entropy Search (MES) from Wang and Jegelka (2017). However, these methods depend on approximations for tractable acquisition optimization, potentially compromising algorithmic soundness. Notably, convergence analysis for constrained multi-objective Bayesian optimization remains unexplored.

# 3 PRELIMINARIES

## 3.1 Problem Statement

Let $[n]$ denote the set $\{1, 2, ..., n\}$ and let $[x]^+$ denote function $\max(0, x)$. For a vector $\mathbf{x}$, its $\ell_2$ norm is denoted by $\|x\|$. For any two vectors $\mathbf{x}_1, \mathbf{x}_2$, we use $\mathbf{x}_1 \leq \mathbf{x}_2$ to denote their element-wise comparisons. To improve the readability of this paper, we utilize the big O notation to omit constant terms in theoretical results.

W.l.o.g., we assume the search space $\mathcal{X}$ is finite, we extend our analysis of continuous and compact search space in Appendix C, e.g. $\mathcal{X} \subseteq [0,1]^d$ is a compact set. Let $F = [f_1, ..., f_m] : \mathcal{X} \to \mathbb{R}^m$ be the multi-objective function with $m$ objective functions. Let $G = [g_1, ..., g_c] : \mathcal{X} \to \mathbb{R}^c$ be the multi-objective function with $c$ constraint functions. Here $f_i, \forall i \in [m], g_j, \forall j \in [c]$ are black-box functions, which implies that they can be non-linear and non-convex functions and we don't necessarily have access to their gradients.

We consider the following constrained multi-objective optimization problem:

$$\max_{\mathbf{x} \in \mathcal{X}} \quad F(\mathbf{x}) = [f_1(\mathbf{x}), ..., f_m(\mathbf{x})],$$
$$\text{s.t.} \quad G(\mathbf{x}) = [g_1(\mathbf{x}), ..., g_c(\mathbf{x})] \geq 0.$$

Since we have $c$ constraints, the feasible region of this problem is given by

$$\mathfrak{F} = \{\mathbf{x} | \mathbf{x} \in \mathcal{X}, g_j(\mathbf{x}) \geq 0, \forall j \in [c]\}. \quad (1)$$

For two data points $\mathbf{x}_1, \mathbf{x}_2 \in \mathfrak{F}$, $\mathbf{x}_1$ is said to *Pareto-dominate* $\mathbf{x}_2$ if (a) $\forall i \in [m], f_i(\mathbf{x}_1) \geq f_i(\mathbf{x}_2)$, and (b) $\exists j \in [m]$ such that $f_j(\mathbf{x}_1) > f_j(\mathbf{x}_2)$. A data point $\mathbf{x}$ is said to be *Pareto-optimal* in $\mathfrak{F}$ if no data point in $\mathfrak{F}$ Pareto-dominates it. Let $\mathcal{X}^*$ denote the set of all Pareto-optimal data points in $\mathfrak{F}$, then the *Pareto front* is defined as $\mathcal{P} = \{F(\mathbf{x}) | \mathbf{x} \in \mathcal{X}^*\}$.

For a compact set $Y$, let $\text{vol}(Y)$ denote its hypervolume. Given a reference point $z$ and a compact set $Y \subseteq F(\mathfrak{F})$, we extend the definition of hypervolume indicator (Golovin and Zhang, 2020) to *constrained hypervolume* as

$\mathcal{HV}_z(Y) =$
$\text{vol}(\{y \in F(\mathfrak{F}) | y \geq z, y \text{ is dominiated by some } y' \in Y\}).$

Then, at step $t \in [T]$, the *simple hypervolume regret* $r_t$ is defined as the difference between the constrained hypervolume indicator of the Pareto front and the current approximation of the Pareto front, i.e.,

$$r_t = \mathcal{HV}_z(\mathcal{P}) - \mathcal{HV}_z(Y_t),$$

where $Y_t$ is the set of observations with $|Y_t| = t$. Maximizing $\mathcal{HV}_z(Y_t)$ reflects the exploration of the Pareto front in the feasible region since it cannot be greater than $\mathcal{HV}_z(\mathcal{P})$. And the *cumulative hypervolume regret* $\mathcal{R}_T$ is defined as the sum of simple hypervolume regret in $T$ iterations, i.e., $\mathcal{R}_T = \sum_{t=1}^T r_t$.

Since we are considering the constrained optimization problem, for a given $\mathbf{x}_t$ at $t$-th iteration, $\forall j \in [c]$, we define *simple violation* as

$$v_{j,t} = [-g_j(\mathbf{x}_t)]^+.$$

Accordingly, $\forall j \in [c]$, the *cumulative violation* after $T$ iterations is defined as $\mathcal{V}_{j,T} = \sum_{t=1}^T v_{j,t}$.

However, hypervolume regret and violations cannot simultaneously assess an algorithm's ability to explore the Pareto front and approximate the feasible region. To address this issue, we define a novel performance metric, called *constraint regret*, generalized from the single objective optimization tasks (Xu et al., 2023), which is the minimum sum of simple hypervolume regret and violations observed up to $t$-th iteration:

$$\mathcal{C}_t = \min_{\tau \in [t]} \left\{ r_\tau + \sum_{j=1}^c v_{j,\tau} \right\}$$

This metric allows for a more comprehensive evaluation of an algorithm's performance in constrained multi-objective settings.

## 3.2 Bayesian Optimization

Bayesian optimization usually runs in a sequential manner. At each iteration $t \in [T]$, the learner takes a new instance $\mathbf{x}_t$, and the observation is generated by

$$y_t = f(\mathbf{x}_t) + \eta_t$$

where $f$ is the objective function, and $\eta_t \sim \mathcal{N}(0, \sigma^2)$ is the observation noise. Let $Y_t = [y_1, ..., y_t]$ be the set of observations and $X_t = [\mathbf{x}_1, ..., \mathbf{x}_t]$ be the set of corresponding historically evaluated candidates. Assume $f$ is drawn from some Gaussian process with $k(\mathbf{x}, \mathbf{x}')$ being the kernel specifying the covariance of $f$ at any two observations $\mathbf{x}, \mathbf{x}'$, i.e., $f \sim \mathcal{GP}(0, k(\cdot, \cdot))$. We further assume $k(\cdot, \cdot) \leq 1$. Consider the posterior predictive distribution of $f(\mathbf{x})$ given $t$ observations: $f(\mathbf{x})|Y_t \sim \mathcal{N}(\mu_t(\mathbf{x}), \sigma_t^2(\mathbf{x}))$ where $\mu_t, \sigma_t$ are specified as:

$$\mu_t(\mathbf{x}) = k(\mathbf{x}_{1:t}, \mathbf{x})^\top (K + \sigma^2 I)^{-1} y_{1:t}, \quad (2)$$
$$\sigma_t^2(\mathbf{x}) = k(\mathbf{x}, \mathbf{x}) - k(\mathbf{x}_{1:t}, \mathbf{x})^\top (K + \sigma^2 I)^{-1} k(\mathbf{x}_{1:t}, \mathbf{x}), \quad (3)$$

where $K = [k(\mathbf{x}, \mathbf{x}')]_{\mathbf{x}, \mathbf{x}' \in \{\mathbf{x}_1, ..., \mathbf{x}_t\}}$ and $k(\mathbf{x}_{1:t}, \mathbf{x}) = [k(\mathbf{x}_1, \mathbf{x}), ..., k(\mathbf{x}_t, \mathbf{x})]^\top$.

Since we have both multiple objective functions and constraint functions, throughout this paper, we assume that they are all drawn from some Gaussian process, i.e., $\forall i \in [m], f_i \sim \mathcal{GP}_i, \forall j \in [c], g_j \sim \mathcal{GP}_j$. Similar to Eq. (2) (3), we can define the posterior mean and variance function for $f_i, g_j$.

Some existing work in Bayesian optimization (Srinivas et al., 2010; Xu et al., 2023; Li and Scarlett, 2022; Zhang et al., 2023a,b) relies on confidence bounds to design provable sample efficient algorithms. Here, we reiterate the construction and guarantee of the confidence bounds, which are essential for the design of a principled constrained multi-objective Bayesian optimization algorithm.

At iteration $t \in [T]$, for function $h \in \{f_i\}_{i \in [m]} \cup \{g_j\}_{j \in [c]}$, the lower confidence bound (LCB) $l_{h,t}$ and upper confidence bound (UCB) $u_{h,t}$ can be defined as:

$$l_{h,t}(\mathbf{x}) := \mu_{h,t-1}(\mathbf{x}) - \beta_t^{\frac{1}{2}} \sigma_{h,t-1}(\mathbf{x}),$$
$$u_{h,t}(\mathbf{x}) := \mu_{h,t-1}(\mathbf{x}) + \beta_t^{\frac{1}{2}} \sigma_{h,t-1}(\mathbf{x}), \quad (4)$$

where $\mu_{h,t-1}$ and $\sigma_{h,t-1}$ are calculated using Eq. (2) (3), and $\beta_t$ is a confidence parameter. With all these UCBs and LCBs at hand, we present Lemma 1 from Srinivas et al. (2010). It shows that with a proper choice of parameters, the confidence interval offers high confidence bound on discrete search space through the whole optimization, where objectives $f_i$ and the constraint functions $g_j$ are bounded by their corresponding UCB and LCB jointly with high probability.

**Lemma 1** (Lemma 5.1, Srinivas et al. (2010))**.** Let $\beta_t = 2\log((m+c)|\mathcal{X}|\pi^2 t^2/6\delta)$. Then, fix $\delta \in (0,1), \forall \mathbf{x} \in \mathcal{X}, \forall t \in [T]$, for function $h \in \{f_i\}_{i \in [m]} \cup \{g_j\}_{j \in [c]}$, with probability $\geq 1 - \delta$,

$$|\mu_{h,t-1}(\mathbf{x}) - h(\mathbf{x})| \leq \beta_t^{1/2} \sigma_{h,t-1}(\mathbf{x}).$$

## 4 METHOD

The challenge of principled constrained multi-objective Bayesian optimization lies in efficiently learning constraints and optimizing within the probably feasible region of multiple objectives. Previous works (Daulton et al., 2020, 2021) typically rely on predicted feasibility or tractable approximations of information criteria (Hernández-Lobato et al., 2016; Fernández-Sánchez et al., 2023) to incorporate constraint learning in optimization. These surrogates and approximations hinder theoretically sound estimation of the constrained hypervolume's behavior during optimization. To address this, we propose generalizing the conventional UCB acquisition function (Chowdhury and Gopalan,

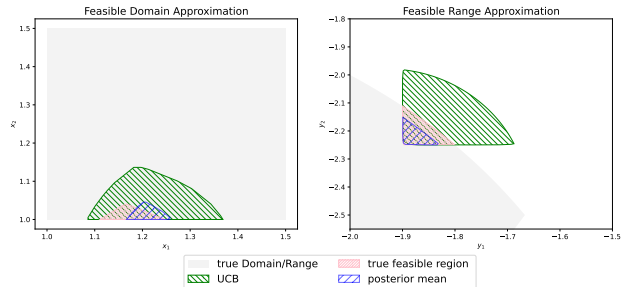

Figure 1: A demonstration of the feasible region estimated by UCBs. The left figure illustrates the estimated feasible domain on the input space $\mathcal{X} \subset \mathbb{R}^2$ of the **Toy Function** ($d = 2$, $m = 2$, $c = 2$), which is discussed in detail in Appendix D. We compare the estimation utilizing the UCBs of the unknown constraints against the estimation using the GPs' posterior means. The right figure displays the area below the feasible Pareto fronts estimated using the UCBs, the posterior means, and the true values. It shows the advantage of applying UCBs in avoiding excluding the global maximum.

2017; Srinivas et al., 2010) for the constrained hypervolume.

We introduce **Co**nstrained **M**ulti-Objective **B**ayesian **O**ptimization through **O**ptimistic Constraints Estimation (**COMBOO**), which efficiently handles unknown objectives and constraints by adopting an optimistic view of these unknowns. At any point, the objective and constraint functions are inferred from their posterior distributions based on current observations. The acquisition function is optimized to guide the search toward promising regions, balancing optimality and feasibility. For optimality, **COMBOO** uses the Upper Confidence Bound (UCB) from single objective Bayesian optimization and employs random scalarization of multiple UCBs for efficient hypervolume optimization. For feasibility, **COMBOO** intersects potentially feasible regions defined by each constraint to avoid querying likely infeasible points. This approach constructs a series of *UCBs for the unbiased Monte Carlo estimator for the constrained hypervolumes through the optimization*. This generalizes the notion of UCB from both an optimality perspective and feasibility perspective and simultaneously takes all the multiple perspectives into account. It allows for explicit search space constraining, principled acquisition, theoretical guarantees, and the ability to declare infeasibility when no feasible candidate exists, as demonstrated in §5.

**Scalarization.** In our multi-objective setting, we address the challenge of trading off multiple acqui-

**Algorithm 1** **Co**nstrained **M**ulti-Objective **B**ayesian **O**ptimization through **O**ptimistic Constraints Estimation (**COMBOO**)

1: **for** $t \in [T]$ **do**
2:    **if** $\max_{\mathbf{x} \in \mathcal{X}} \{\min_{j \in [c]} u_{g_j,t}(\mathbf{x})\} < 0$ **then**
3:       **Declare infeasibility.**
4:    **end if**
5:    Sample $\theta_t$ uniformly from $\mathcal{S}^+_{k-1}$ as in (5)
6:    Find $\mathbf{x}_t \in \arg\max_{\mathbf{x} \in \mathcal{X}} \alpha_t(\mathbf{x})$ as in (7)
7:       s.t. $u_{g_j,t} \geq 0, \forall j \in [c]$ as in (4).
8:    Evaluate $F, G$ at $\mathbf{x}_t$.
9:    Update GP posteriors with new evaluations.
10: **end for**

sition functions by applying a scalarization mapping $s_{\theta_t}$ as defined in (5), parameterized by a randomly drawn variable $\theta_t$ at each iteration. This *hypervolume scalarization*, introduced by Deng and Zhang (2019); Golovin and Zhang (2020), allows for the Monte Carlo estimator of the hypervolume and its estimation error. This approach enables a principled combination of optimizing objectives and considering unknown constraints. We extend this scalarization to the constrained optimization scenario with both theoretical guarantees and comprehensive empirical evidence of its efficiency. Here, we define the acquisition function in Algorithm 1 with scalarization of UCBs.

**Definition 4.1** (Scalarization function (Deng and Zhang, 2019; Golovin and Zhang, 2020)). The *hypervolume scalarization* is defined as

$$s_\theta(y) = \min_{i \in [m]} ([y_i/\theta_i]^+)^m \text{ s.t. } y, \theta \in \mathbb{R}^m. \quad (5)$$

Furthermore, it holds that

$$\mathcal{HV}_z(Y_t) = c_m \mathbb{E}_{\theta \sim \mathcal{S}^+_{m-1}} [\max_{y \in Y_t \cap F(\mathfrak{F})} s_\theta(y - z)] \quad (6)$$

where $\theta \sim \mathcal{S}^+_{m-1}$ denote drawing $\theta$ uniformly from $\mathcal{S}^+_{m-1} = \{y \in \mathbb{R}^m | \|y\| = 1, y \geq 0\}$ and $c_m = \frac{\pi^{\frac{m}{2}}}{2^m \Gamma(\frac{m}{2}+1)}$.

This scalarization function is derived from the integration in the calculation of hypervolume, offering an unbiased estimation of the HV. When applying a sample of the random scalarization to the UCBs of the objectives at a certain time $t$, we have the following acquisition function.

**Definition 4.2** (Acquisition function). Let the $m$-dimensional vector $U_t(\mathbf{x})$ denote the UCB of $m$ objectives $f_i$. We define the acquisition function $\alpha_t(\mathbf{x})$ as the value of $U_t(\mathbf{x})$ scalarized by *hypervolume scalarization* as defined in Eq. (5).

$$U_t(\mathbf{x}) = [u_{f_1,t}(\mathbf{x}) - z_1, ..., u_{f_m,t}(\mathbf{x}) - z_m]$$
$$\alpha_t(\mathbf{x}) = s_{\theta_t}(U_t(\mathbf{x})) \quad (7)$$

where $z = (z_1, ..., z_m)$ is a chosen sub-optimal value.

Now, we need to construct an optimistic estimation of the constraint functions to incorporate the consideration of the feasibility.

**Optimistic estimation of feasibility.** Recent work (Xu et al., 2023; Zhang et al., 2023b) in constrained single-objective Bayesian optimization inspired **COMBOO** by using the UCB of unknown constraints to limit the search space. Specifically, **COMBOO** incorporates feasibility through upper confidence bounds $w_j$ for the $c$ constraint functions $g_j$ as defined in Eq. (4). When optimizing the scalarized acquisition function, **COMBOO** ensures the search space is constrained so all UCBs for the constraints exceed the corresponding thresholds, shown in line 7 of *Algorithm 1*. This guarantees objectives are optimized in a likely feasible region as is shown in Figure 1. Although the previous work adopts a frequentist view rather than our Bayesian framework based on the Gaussian process and only focuses on a single objective rather than the Pareto front, we generalize its results to achieve the guarantees on cumulative violation for **COMBOO**, thus bridging multi-objective BO and constrained single-objective BO.

**Constrained optimization.** With the optimistic estimation of feasibility discussed above and the sample from the random scalarization adaptive tradeoff among multiple objectives, we can define the **COMBOO** optimization loop. In each iteration, we maximize the scalarized function subject to the newly defined constraints. In line 3 of Algorithm 1, we solve an auxiliary optimization problem to determine whether infeasibility should be declared. The solution to this auxiliary problem, $\arg\max_{\mathbf{x} \in \mathcal{X}} \min_{j \in [c]} u_{g_j,t}(\mathbf{x})$, can also be leveraged in the optimization of the acquisition function, as it helps discard inactive constraints. Note that the UCB of $Y_t$ in the unconstrained hypervolume shown in Eq. (6) at time $t$ is upper bounded by the maximum of the expectation with respect to $\mathcal{S}^+_{m-1}$ of the acquisition function defined in Eq. (7), detailed explanation of which is deferred to Appendix B. Combined with the optimistic feasibility estimation in line 4 of Algorithm 1, we know that **COMBOO** iteratively picks the maximizer of the *UCB of the Monte Carlo estimator of the constrained hypervolume*. This allows the following theoretical guarantee of **COMBOO**.

## 5 THEORETICAL RESULTS

We analyze the performance of Algorithm 1 for discrete search spaces here. The omitted proofs of our results are provided in Appendix B. Additionally, in Appendix C, we extend this analysis to continuous and compact search spaces, following a similar approach and requiring only an appropriate adjustment of the confidence parameter $\beta_t$.

## 5.1 Results under Feasibility Assumption

Recall that $\mathfrak{F}$ denotes the set of all feasible solutions defined in Eq. (1). In this subsection, we discuss the case when the problem is feasible, i.e., $\mathfrak{F} \neq \emptyset$. We first provide the bounds for cumulative HV regret and cumulative violation regret (Theorem 1, Theorem 2), followed by a high probability bound on the false positive rate in Line 3 of Algorithm 1 (Theorem 3).

**Hypervolume regret bound.** The bound for cumulative HV regret is provided as follows.

**Theorem 1** (Cumulative HV regret bound). *With the conditions in Lemma 1, $\forall T \geq 1$, the following holds with probability at least $1 - \delta$*

$$\mathcal{R}_T \leq O(m^2[\gamma_T T \ln(T)]^{1/2})$$

*Remark.* The cumulative HV bound is dependent on the Lipschitz constant of the chosen scalarization function $s_\theta$ and the maximum information gain (MIG), $\gamma_T$ of the kernel used in the GP model, as defined in Srinivas et al. (2010). The orders of MIGs for common kernels were also shown in this work. The proof of Theorem 1 requires a novel adaptation from the regret analysis of Golovin and Zhang (2020) and Paria et al. (2020) to the Bayesian setting. Furthermore, instead of a deterministic bound as originally shown in Golovin and Zhang (2020), we need to establish a high-probability bound due to the task of constraint learning.

Additionally, from the high-probability bound on the cumulative HV regret derived above, it is straightforward to obtain an upper bound for the simple HV regret $r_T$ as $r_T \leq O\left(m^2 \sqrt{\frac{\gamma_T \ln T}{T}}\right)$.

**Cumulative violation bound.** We have a similar bound for cumulative violation.

**Theorem 2** (Cumulative constraint violation bound). *$\forall j \in [c]$, with conditions of Lemma 1,*

$$\mathcal{V}_{j,T} \leq O((\gamma_T T \ln T)^{1/2})$$

*with at least $1 - \delta$ probability.*

*Remark.* It is imperative to recognize that in Theorem 2, the bound for $\mathcal{V}_T$ fundamentally represents the cumulative constraint violation bound as delineated in Xu et al. (2023), where $\beta_T^{1/2} = O(\sqrt{\ln T})$ is substituted for $O(\sqrt{\gamma_T})$ from prior studies, reflecting divergent assumptions regarding the nature of the unknown underlying functions.

**Constraint regret bound.** Furthermore, combining Theorem 1 and Theorem 2, we can derive the convergence rate of constraint regret $\mathcal{C}_T$, defined in §3.1.

**Corollary 1** (Convergence of constraint regret). *$\forall T \geq 1$, with probability at least $1 - \delta$*

$$\mathcal{C}_T \leq O((c + m^2)[\gamma_T \ln(T)/T]^{1/2})$$

**Guarantee for low false positive rate.** Under the regular assumption that a feasible solution exists, it is possible—due to the stochastic nature of our algorithm and the randomness of initial sampling—that Line 3 of Algorithm 1 may return a negative value, leading to a false declaration of infeasibility and termination of the algorithm. However, we can demonstrate that the algorithm **COMBOO**, will avoid this type of misjudgment with high probability under the regular assumption.

**Theorem 3** (Declaration of infeasibility in feasible case). *With conditions in Lemma 1, if the problem is feasible, i.e.*

$$\exists \mathbf{x} \in \mathcal{X}, \forall j \in [c], g_j(\mathbf{x}) \geq 0$$

*then, Algorithm 1 will not declare infeasibility with probability $\geq 1 - \delta$.*

## 5.2 Results under Infeasibility Assumption

For completeness, in this subsection, we discuss the case when $\mathfrak{F} = \emptyset$, i.e., in extremely rare instances, the optimization problem may be infeasible. We demonstrate that, in such cases, the algorithm will identify and declare infeasibility within a finite number of steps, with high probability (Theorem 4). This guarantees that the algorithm can efficiently detect infeasibility and prevent unnecessary computational effort.

**Theorem 4** (Declaration of infeasibility when the problem is infeasible). *With the conditions in Lemma 1, and that $\lim_{T \to \infty} \frac{\beta_T^{\frac{1}{2}} \sqrt{\gamma_T}}{\sqrt{T}} = 0$. If the problem is infeasible, i.e.*

$$\forall \mathbf{x} \in \mathcal{X}, \min_j g_j(\mathbf{x}) < 0.$$

*Then, given $\delta \in (0, 1)$, with probability at least $1 - \delta$, Algorithm 1 will declare infeasibility within the following number of steps:*

$$\bar{T} = \min_{T \in \mathbb{N}^+} \left\{ T \middle| \frac{\beta_T^{1/2} \sqrt{\gamma_T}}{\sqrt{T}} \leq \frac{\epsilon}{C} \right\},$$

*where $C$ is a positive constant independent of $T$ and $\epsilon = |\max_{x \in \mathcal{X}} \min_{j \in [c]} g_j(x)|$.*

*Remark.* The extra condition $\lim_{T\to\infty} \frac{\beta_T^{\frac{1}{2}}\sqrt{\gamma_T}}{\sqrt{T}} = 0$ depends on the choice of kernel and $\beta_t$, which holds with our choice of $\beta_t$ combined with Linear or RBF kernel. For the Matérn kernel, it must hold that $v > d/2$, where $v, d$ are kernel parameters (Appendix D).

# 6 EXPERIMENTS

We applied **COMBOO** to various synthetic test functions and datasets, including **Toy Function** ($d = 2, m = 2, c = 2$), **Branin-Currin** ($d = 2, m = 2, c = 2$), and **C2-DTLZ2** ($d = 4, m = 2, c = 1$) as discussed in details in Appendix D. Experiments conducted on real-world application problems include **Penicillin Function** ($d = 7, m = 3, c = 3$) (Liang and Lai, 2021), **Disc Brake Design Problem** (d = 4, m = 2, c= 3) (Tanabe and Ishibuchi, 2020), as well as discrete problems **Caco-2++** ($d = 2175, m = 3, c = 3$) adapted from (Park et al., 2024) and **ESOL+**($d = 2133, m = 4, c = 4$) adapted from Delaney (2004).

We compared **COMBOO** with various benchmarks, including Parallel Noisy Expected Hypervolume Improvement (**qNEHVI**) (Daulton et al., 2021), parallel ParEGO (**qParEGO**) (Daulton et al., 2020), Max-value Entropy Search for Multi-Objective Bayesian Optimization with Constraints (**MESMOC**) (Belakaria et al., 2020a), and Random Search. All benchmarks except **MESMOC** were implemented using the Python library BoTorch (Balandat et al., 2020). Detailed experimental settings and full results are given in Appendix D. To balance the contributions of HV regret and violation of constraints, we normalize $r_\tau$ and $\sum_{j=1}^c v_{j,\tau}$ before measuring $\mathcal{C}_t$. We used 1.96 standard error from 10 independent trials to construct the shaded areas for all the curves.

**To match the parallel settings**. Though the benchmark algorithms were designed for batched output, we take a number of queries $q = 1$ in each step to make them comparable to our approach, which is also seen in the experiments in Daulton et al. (2021).

**Acquisition on discrete and continuous space.** In this study, we avoided manually discretizing the domain for **COMBOO** whenever a continuous search space and objective evaluation were available. Instead, we used a multi-start gradient descent optimizer for the acquisition function. Although we introduced discretization in §5 for theoretical convenience, it is not a fundamental limitation of the algorithm, as demonstrated in Appendix C. Moreover, given sufficient computational resources, the discretization granularity can be made arbitrarily fine, effectively closing the gap with fully continuous optimization. In summary, our approach provides a consistent and effective solution across various search spaces.

## 6.1 Constraint Handling across Various Benchmarks

**qNEHVI** (Daulton et al., 2021) primarily aimed at unconstrained multi-objective optimization and offered the derivation handling noisy outcome constraints. It calculates the expectation of $\mathcal{HV}_z(Y_t)$ with respect to $F(\mathbf{x}_t)$, formulated as $E_{F(\mathbf{x}_t)}[\mathcal{HV}_z(Y_t)]$. It addresses constraints by employing a conditional expectation:

$$\mathbf{x}_t = \arg\max E_{F(\mathbf{x})}[\mathcal{HV}_z(Y_t)|\mathbf{x} \in \mathfrak{F}].$$

**qParEGO** (Daulton et al., 2020) utilizes a random augmented Chebychev scalarization $s_{\theta_t}(y) = \min_{i\in[m]} \theta_{i,t}(y_i - z_i)$ at each step for the objectives and employs GP to model the scalarized outcomes (Knowles, 2006), represented as $\bar{s}_{\theta_t}(\mathbf{x})$. Subsequently, it applies conditional expected improvement to the surrogate scalarized objective.

$$\mathbf{x}_t = \arg\max E_{F(\mathbf{x})_{\theta_t}}\left[[\bar{s}_{\theta_t}(\mathbf{x}) - s_{\theta_t}^*]^+ \times \mathbb{I}(\mathbf{x} \in \mathfrak{F})\right]$$

where $\mathbb{I}(\mathbf{x})$ is an indicator for $\mathbf{x}$ being feasible, also approximated by surrogates of $g_j$, and $s_{\theta_t}^*$ is the best observation of scalarized objective in the current step.

**MESMOC** also handles unknown constraint functions by modeling each as an independent GP surrogate trained on past evaluations. For each candidate input, constraints are sampled from their GPs and integrated into a cheap multi-objective optimization (via NSGA-II (Deb et al., 2002)) to generate Pareto front samples that satisfy the approximated constraints.

## 6.2 Results on Real-world Problems

We introduce a real-world drug discovery problem: **Caco-2++**, a dataset of 906 molecules adapted from **Caco-2** (Wang et al., 2016), sourced from the Therapeutics Data Commons (Huang et al., 2021). The search space includes molecular fingerprints and fragments (*fragprint*) (Griffiths et al., 2022), along with the *mqn feature*. In addition to permeability contained in the original dataset, which assesses the ADME profile of drugs (Park et al., 2024), we added TPSA and drug-likeness (QED) (Bickerton et al., 2012) as objectives for a CMO problem. Constraints are lower thresholds for each objective, and these properties cannot be optimized simultaneously. The Tanimoto kernel (Gower, 1971) was employed in the GP models, whose implementation was facilitated using the Python library GAUCHE (Griffiths et al., 2023).

In Figure 4, we plot the observed best HV and cumulative constraint violation for the benchmarks on the **Caco-2++** dataset over time. The scalarization-based methods, **COMBOO** and **qParEGO**, showed faster convergence in simple HV compared to non-scalarization approaches like **qNEHVI** and **MESMOC**.

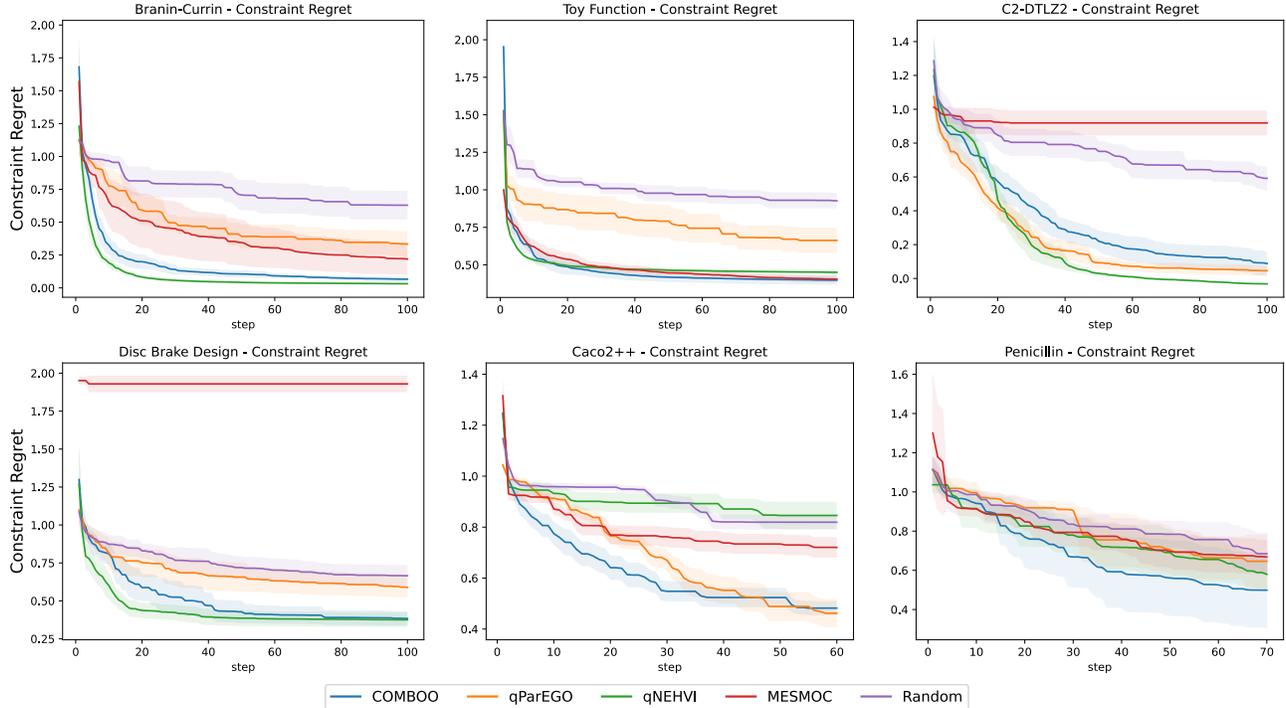

Figure 2: The constraint regret of benchmarks on synthetic and real-world problems. The first row shows the results of synthetic problems, while the second row shows the results of real-world problems. We report the results from 10 independent trials on each benchmark and algorithm.

We also evaluated the efficiency of constraint handling across the various methods. Notably, the information-theoretic method, **MESMOC**, exhibited near-zero constraint violation throughout the trials, in contrast to other methods, including **COMBOO**, which showed increasing violations at different rates. This behavior can be attributed to **MESMOC**'s more pessimistic constraint exploration strategy.

**MESMOC** utilizes the posterior samples from the GP model, $\mu_{h,t}$, to enforce constraints during the optimization of the acquisition function, as outlined in Line 7 of Algorithm 1. In contrast, our method employs UCB, $u_{h,t}$, to optimistically search for feasible regions, thereby expanding the search space for acquisition function optimization and increasing the likelihood of identifying optimal solutions. While the optimistic approximation may lead to more frequent constraint violations, it simultaneously facilitates a more comprehensive exploration of objective values, resulting in a favorable trade-off between constraint violation and HV regret.

We measure this trade-off by the constraint regret, $\mathcal{C}_t$.

From Figure 2, we found **COMBOO** outperforms or is comparable to all benchmarks in terms of constraint regret on real-world problems.

### 6.3 Results on Synthetic Problems

In this section, we evaluate the performance of various benchmarks on synthetic test functions. We conducted experiments on a customized **Toy Function**, as well as two widely used CMO test functions: **Branin-Currin** and **C2-DTLZ2**. The constraint regret curves for these functions are presented in Figure 2.

In the series of experiments conducted, **COMBOO** displayed a consistently strong performance, often aligning with the top-performing methods across various test functions. Conversely, **MESMOC** showed the least stable results, particularly when dealing with more intricate constraints, as observed in **C2-DTLZ2**. Meanwhile, **qNEHVI** demonstrated quicker convergence on smaller-scale test functions within this experimental framework; however, its performance did not sustain this lead when applied to larger-scale scenarios, such as the **Penicillin Function** in real-world applications.

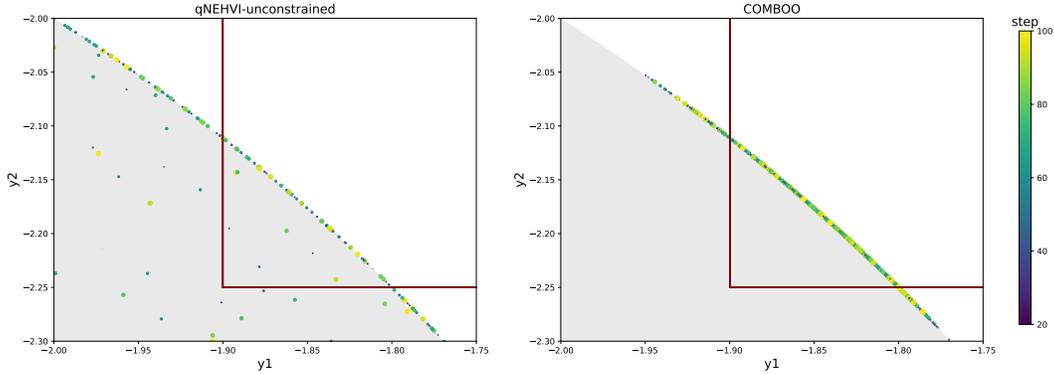

Figure 3: The objective values queried by the unconstrained version of **qNEHVI** (left) and **COMBOO** (right) over 10 trials. The feasible region defined by $y_1 \geq -1.9$ and $y_2 \geq -2.25$, lies above the maroon boundary lines. The range of this problem is shaded by gray area, from which we can observe the whole Pareto front. The order in which each point was queried is indicated by shades of colors and increasing point size.

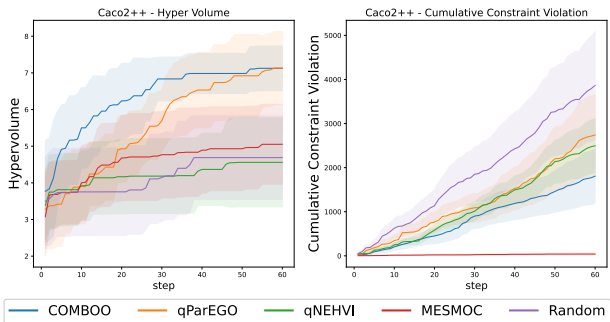

Figure 4: The hypervolume (left) and cumulative constraints violation (right) on the **Caco-2++** dataset.

### 6.4 Efficiency of Constraint Learning

A motivating problem is optimizing multiple objectives in experimental settings, such as penicillin production, where objectives must meet thresholds $f_i(\mathbf{x}) \geq s_i$ to satisfy regulatory requirements. Unlike general constraints $g_j(\mathbf{x}) \geq 0$, these constraints are the objectives themselves. As a result, the Pareto front is a subset of the unconstrained solution. To assess if **COMBOO** effectively prioritizes the Pareto front within the feasible region, we compared it with the state-of-the-art algorithm **qNEHVI** under unconstrained settings.

Methods specialized for CMO outperformed others in terms of simple HV regret and cumulative violation for both the **Toy function** and **Penicillin function** (Figure 5). We also present the sample trajectory of the two-dimensional **Toy function** (Figure 3). **COMBOO** not only concentrated on the feasible region as indicated by the red lines, but also kept the infeasible observations in the global Pareto front. In contrast, the unconstrained algorithm wasted observations in regions distant from the Pareto front.

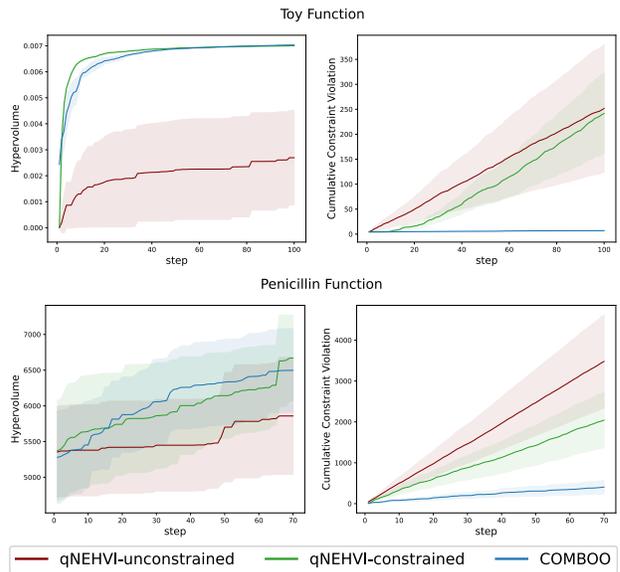

Figure 5: The hypervolume (left) and cumulative constraints violation (right) on the **Toy** and **Penicillin** functions. We compare **COMBOO** with both the constrained and unconstrained version of **qNEHVI**.

## 7 CONCLUSION

We proposed **COMBOO**, a stochastic scalarization-based Bayesian optimization algorithm utilizing optimistic estimations for constrained multi-objective optimization. By using random scalarization to connect the UCBs of multiple objectives and optimistic feasible region estimation based on the UCBs for the constraints, we construct the optimistic estimation for the constrained hypervolume, which incorporates both optimality and feasibility for **COMBOO**. We demonstrate the convergence theoretically, with empirical results showing its robustness across various benchmarks.

# Acknowledgments

This work was partially done when CL was at the University of Chicago. This work was supported in part by the National Science Foundation under Grant No. IIS 2313131, IIS 2332475 and CMMI 2037026. The authors acknowledge the University of Chicago's Research Computing Center for their support.

## A SUPPLEMENTAL DEFINITIONS

In this section, we provide the formal definition of MIG $\gamma_T$ used in §5. Moreover, we define the four additional values that were not empirically measured but are used in the following sections of theoretical analysis.

**Definition A.1** (Maximum information gain). The information gain of a function $f$ is the mutual information of the distribution of $f$ and the distribution of $t$ function observations $Y_t$, denoted as $I(Y_t; f)$. Moreover, let the set of instances corresponding to each element in $Y_t$ be $A_t$. Both $f$ and $Y_t$ follow the assumption for the GP model in §3.2.

$$I(Y_t; f) = H(f) - H(f|Y_t) = H(Y_t) - H(Y_t|f) \quad (8)$$

where $H(\cdot)$ is the Shannon entropy function. The closed form of $I(Y_t; f)$ is given by Srinivas et al. (2012): $\frac{1}{2} \log \det(I + \lambda^{-1} K_{i,t})$ where $K_t = [k(\mathbf{x}, \mathbf{x}')_{\mathbf{x}, \mathbf{x}' \in A_t}]$. Accordingly, the maximum information gain (MIG) for objective $f$ given $t$ observations is defined as:

$$\gamma_t = \max_{A_t \subset \mathcal{X} \text{ s.t. } |A_t|=t} \frac{1}{2} \log \det(I + \sigma^{-2} K_t) \quad (9)$$

We denote the MIG of a function $h \in \{f_i\}_{i \in [m]} \cup \{g_j\}_{j \in [c]}$ given the current $T$ observations, as $\gamma_{h,T}$. Let $\gamma_T$ be an upper bound of $\gamma_{h,T}, \forall h \in \{f_i\}_{i \in [m]} \cup \{g_j\}_{j \in [c]}$.

**Definition A.2** (Scalarized regret). We define the scalarized regret as

$$r_\theta(X_t) = \max_{\mathbf{x} \in \mathfrak{F}} s_\theta(F(\mathbf{x}) - z) - \max_{\mathbf{x} \in X_t \cap \mathfrak{F}} s_\theta(F(\mathbf{x}) - z) \quad (10)$$

**Definition A.3** (Bayesian regret). The Bayesian regret is the expectation of scalarized regret over all possible $\theta \sim \mathcal{S}_{m-1}^+$:

$$R(X_t) = E_\theta[r_\theta(X_t)] \quad (11)$$

**Definition A.4** (Instantaneous regret). The instantaneous regret is defined as:

$$r'(\mathbf{x}_t, \theta_t) = \max_{\mathbf{x} \in \mathfrak{F}} s_{\theta_t}(F(\mathbf{x}) - z) - s_{\theta_t}(F(\mathbf{x}_t) - z) \quad (12)$$

where $\theta_t$ and $\mathbf{x}_t$ are the parameter and corresponding observation in the $t^{th}$ iteration of our algorithm.

**Definition A.5** (Cumulative regret). The cumulative regret is defined as the sum of instantaneous regret:

$$R_C(T) = \sum_{t=1}^T r'(\mathbf{x}_t, \theta_t) \quad (13)$$

## B PROOFS FOR FINITE DISCRETE SEARCH SPACE

### B.1 Proof of Theorem 1

We first state the necessary lemmas for the proof.

**Lemma 2** (First part of Lemma 6 in Golovin and Zhang (2020)). For

$$s_\theta(y) = \min_{i \in [m]} ([y_i/\theta_i]^+)^m \text{ s.t. } y, \theta \in \mathbb{R}^m$$

in (5), it is $O(B^m m^{1+m/2})$-Lipschitz for all $\theta$, where $B \geq y - z$, $z$ is the chosen sub-optimal point.

**Lemma 3** (Lemma 5 in Golovin and Zhang (2020)). (6) holds. That is,

$$\mathcal{HV}_z(Y_t) = c_m \mathbb{E}_{\theta \sim \mathcal{S}_{m-1}^+}[\max_{y \in Y_t \cap F(\mathfrak{F})} s_\theta(y - z)]$$

**Lemma 4** (Modified version of Theorem 7 in Golovin and Zhang (2020)). Suppose $s_\theta(y)$ is L-Lipschitz for all possible $\theta \in \mathcal{S}_{m-1}^+$ (that is, $|s_\theta(y) - s_\theta(y')| \leq L\|y - y'\|_1$), with GP models following the definition in §3.1, then the expected cumulative regret in Definition 13 is bounded with probability at least $1 - \delta$

$$E[R_C(T)] = O(Lm[\gamma_T T \ln(T)]^{1/2}) \quad (14)$$

*Proof of Lemma 4.* The unconstrained and definite version of this lemma was directly used in Golovin and Zhang (2020) and was proved by Paria et al. (2020). Define $\mathfrak{F}_t = \{\mathbf{x} \mid u_{g_j,t}(\mathbf{x}) \geq 0, \forall j \in [c]\}$. Let's pick $\mathbf{x}_t^* = \arg\max_{\mathbf{x} \in \mathfrak{F}_t} s_{\theta_t}(F(\mathbf{x}))$ and $\mathbf{x}_t = \arg\max_{\mathbf{x} \in \mathfrak{F}_t} s_{\theta_t}(U_t(\mathbf{x}))$ in our problem. Then it follows that

$$E[R_C(T)] = E\left[\sum_{t=1}^T \left(\max_{\mathbf{x} \in \mathfrak{F}} s_{\theta_t}(F(\mathbf{x})) - s_{\theta_t}(F(\mathbf{x}_t))\right)\right]$$

$$\leq \underbrace{E\left[\sum_{t=1}^T s_{\theta_t}(U_t(\mathbf{x}_t)) - s_{\theta_t}(F(\mathbf{x}_t))\right]}_{B_1} +$$

$$\underbrace{E\left[\sum_{t=1}^T s_{\theta_t}(F(\mathbf{x}_t^*)) - s_{\theta_t}(U_t(\mathbf{x}_t^*))\right]}_{B_2} \quad (15)$$

The inequality (15) holds if

$$s_{\theta_t}(U_t(\mathbf{x}_t)) - s_{\theta_t}(U_t(\mathbf{x}_t^*)) \geq 0$$

and

$$s_{\theta_t}(F(\mathbf{x}_t^*)) - \max_{\mathbf{x} \in \mathfrak{F}} s_{\theta_t}(F(\mathbf{x})) \geq 0$$

The first condition holds due to the choice of $\mathbf{x}_t$. From Lemma 1, we know $w_{g_j,t}(\mathbf{x}) \geq g_j(\mathbf{x}), \forall j \in [c], t \in [T], \mathbf{x} \in \mathcal{X}$ with probability at least $1 - \delta$. Hence, we have $\mathfrak{F} \subset \mathfrak{F}_t$ with probability at least $1 - \delta$, meaning the second condition holds with the same probability.

For finite $\mathcal{X}$, the two terms in (15) have the following bounds. Lemma 3 in Paria et al. (2020) shows

$$B_1 \leq \left[L(m\beta_T T)^{1/2} \left(\sum_{i=1}^m \frac{\gamma_{i,T}}{\ln(1 + \sigma^{-2})}\right)^{1/2}\right] + Lm\frac{\delta}{(m+c)\sqrt{2\pi}|\mathcal{X}|} \quad (16)$$

where $\beta_T = \beta_{i,T}, \forall i \in [m]$. Lemma 2 in Paria et al. (2020) shows

$$B_2 \leq Lm \sum_{t=1}^T \sum_{\mathbf{x} \in \mathcal{X}} \exp(-\frac{\beta_{i,t}}{2})$$

$$\leq Lm\frac{\delta}{(m+c)\sqrt{2\pi}} \quad (17)$$

The final bound comes from the fact that $\beta_{i,t} \leq O(\ln t)$ and $\exp(\frac{-\beta_{i,t}}{2}) \leq O(\frac{1}{t^2})$ and that $\sum_{t=1}^T \frac{1}{t^2} \leq \frac{\pi^2}{6}$. $\square$

Now, we are ready to use the relationships between instantaneous regret, cumulative regret, and Bayesian regret, along with the aforementioned lemmas, to derive an upper bound for the cumulative HV regret.

*Proof of Theorem 1.* Consider the newly introduced definitions in Appendix A. We utilize the bound of cumulative regret $R_C(T)$ in Definition 13, which is provided by Lemma 4, to derive a similar bound for $\mathcal{R}_t$. The numerical connection between $R_C(T)$ and $\mathcal{R}_t$ is given by Lemma 3.

From Lemma 3:

$$r_t = \mathcal{HV}_z(\mathcal{P}) - \mathcal{HV}_z(Y_t) = c_m R(X_t) \quad (18)$$

Assume the problem is feasible and infeasibility is not declared, and w.l.o.g. we assume reference point $z = (0, ..., 0)$. The arithmetic relationship shows

$$R(X_t) = E_{\theta_t}[\max_{\mathbf{x} \in \mathfrak{F}} s_{\theta_t}(F(\mathbf{x})) - \max_{x \in X_t \cap \mathfrak{F}} s_{\theta_t}(F(\mathbf{x}))] \quad (19)$$

$$\leq E_{\theta_t}[\max_{\mathbf{x} \in \mathfrak{F}} s_{\theta_t}(F(\mathbf{x})) - s_{\theta_t}(F(\mathbf{x}_t))] \quad (20)$$

$$\leq E_{\theta_t}[r'(\mathbf{x}_t, \theta_t)] \quad (21)$$

Lemma 2 shows $c_m L \leq O(m)$ for $s_\theta$. Combining with Lemma 4, we obtain

$$\mathcal{R}_T = \sum_{t=1}^T c_m R(X_t)$$

$$\leq c_m \sum_{t=1}^T E_{\theta_t}[r'(\mathbf{x}_t, \theta_t)]$$

$$= c_m E[R_C(T)]$$

$$\leq O(m^2[\gamma_T T \ln(T)]^{1/2})$$

with probability at least $1 - \delta$. □

**Remark.** Note that in (20), the same relationship still holds for $U_t$: $\max_{x \in X_t \cap \mathfrak{F}} s_{\theta_t}(U_t(\mathbf{x})) \geq s_{\theta_t}(U_t(\mathbf{x}_t))$. Adding with the expectation with respect to $\theta_t$, the scalarization representation of hypervolume of UCBs of the current objective upper bounds the maximum of acquisition function $\alpha_t(\mathbf{x}_t)$ as described in §4.

### B.2 Proof of Theorem 2

We first state the following lemma bounds the simple violations and variance funciton.

**Lemma 5.** With the conditions in Lemma 1, the following inequality holds,

$$v_{j,t} \leq 2\beta_{j,t}^{1/2} \sigma_{j,t-1}(\mathbf{x}_t), \ \forall j \in [c], \forall t \in [T] \quad (22)$$

with probability at least $1 - \delta$.

*Proof of Lemma 5.* With the definition of $\mathbf{x}_t$ in Algorithm 1, and the fact

$$-[a+b]^- \leq -[a]^- - [b]^- \text{ and } [-x]^+ = -[x]^- \quad (23)$$

we have $\forall j \in [c]$,

$$\begin{aligned}
v_{j,t} &= [-g_j(\mathbf{x}_t)]^+ = -[g_j(\mathbf{x}_t)]^- \\
&= -[g_j(\mathbf{x}_t) + u_{g_j,t}(\mathbf{x}_t) - u_{g_j,t}(\mathbf{x}_t)]^- \\
&\leq -[g_j(\mathbf{x}_t) - u_{g_j,t}(\mathbf{x}_t)]^- - [u_{g_j,t}(\mathbf{x}_t)]^- \\
&= -[g_j(\mathbf{x}_t) - u_{g_j,t}(\mathbf{x}_t)]^- \quad (24) \\
&\leq -[l_{g_j,t}(\mathbf{x}_t) - u_{g_j,t}(\mathbf{x}_t)]^- \\
&= 2\beta_T^{\frac{1}{2}} \sigma_{j,t-1}(\mathbf{x}_t)
\end{aligned}$$

with probability $\geq 1 - \delta$. $[\cdot]^-$ is defined as $\min(0, \cdot)$. The equation in (24) is true by the feasiblibilty of $\{\mathbf{x} \mid u_{g_j,t}(\mathbf{x}) \geq 0, \forall j \in [c]\}$, which holds with at least $1 - \delta$ probability by Theorem 3. □

**Lemma 6** (Lemma 4 in Chowdhury and Gopalan (2017)). $\forall j \in [c]$, with $\mathbf{x}_1, ..., \mathbf{x}_T$ selected by our algorithm,

$$\sum_{t=1}^{T} \sigma_{g_j,t-1}(\mathbf{x}_t) \leq \sqrt{4(T+2)\gamma_{g_j,T}}$$

Now, we are ready to prove the theorem.

*Proof of Theorem 2.*

$$\begin{aligned}
\mathcal{V}_{j,T} &= \sum_{t=1}^{T} v_{j,t} \\
&\leq \sum_{t=1}^{T} 2\beta_t^{1/2} \sigma_{g_j,t-1}(\mathbf{x}_t) \quad (25) \\
&\leq 2\beta_T^{1/2} \sum_{t=1}^{T} \sigma_{g_j,t-1}(\mathbf{x}_t) \quad (26) \\
&\leq 4\beta_T^{1/2} \sqrt{(T+2)\gamma_{g_j,T}} \quad (27) \\
&\leq 4\beta_T^{1/2} \sqrt{(T+2)\gamma_T}
\end{aligned}$$

The inequality in (25) follows directly from Lemma 5. The inequality in (26) comes from the monotonicity of $\beta_t$ with respect to $t$. The inequality (27) follows from Lemma 6. □

### B.3 Proof of Theorem 3

*Proof of Theorem 3.* Until the $T^{th}$ iteration, given $\delta \in (0,1)$ from Lemma 1, we conclude that the following holds with probability at least $1 - \delta$

$$u_{g_j,t}(\mathbf{x}) \geq g_j(\mathbf{x})$$

$$\forall \mathbf{x} \in \mathcal{X}, \forall t \in [T], \forall j \in [c].$$

Then, assuming feasibility, it follows that

$$\max_{\mathbf{x} \in \mathcal{X}} \min_{j \in [c]} u_{g_j,t}(\mathbf{x}) \geq \max_{\mathbf{x} \in \mathcal{X}} \min_{j \in [c]} g_{j,t}(\mathbf{x}) \geq 0 \quad (28)$$

with probability at least $1 - \delta$, which is exactly the counter statement for the condition in line 2 of Algorithm 1. □

### B.4 Proof of Theorem 4

*Proof of Theorem 4.* Let's denote $|\max_{\mathbf{x} \in \mathcal{X}} \min_{j \in [c]} g_j(\mathbf{x})|$ by $\epsilon$.

By a similar conclusion in Lemma 6, we have

$$\forall j \in [c], \ u_{g_j,T}(\mathbf{x}) - l_{g_j,T}(\mathbf{x}) < C \frac{\beta_T^{1/2} \sqrt{\gamma_T}}{\sqrt{T}} \quad (29)$$

for some positive constant $C$. By the regularity condition that $\frac{\beta_t^{1/2} \sqrt{\gamma_t}}{\sqrt{t}} \to 0$, then for sufficient large $T$,

$$u_{g_j,T}(\mathbf{x}) - l_{g_j,T}(\mathbf{x}) < \epsilon \quad (30)$$

Jointly with Lemma 1 for LCB of the constraint function, we have

$$u_{g_j,T}(\mathbf{x}) - g_j(\mathbf{x}) < \epsilon, \forall \mathbf{x} \in \mathcal{X}, \forall j \in [c] \quad (31)$$

which means, for sufficient large $T$,

$$\max_{\mathbf{x} \in \mathcal{X}} \min_{j \in [c]} u_{g_j, T}(\mathbf{x}) < \epsilon - \epsilon = 0 \quad (32)$$

with probability at least $1 - \delta$. This satisfies the rejection condition in line 2 of Algorithm 1. □

### B.5 Proof of Corollary 1

*Proof of Corollary 1.* Since

$$\sum_{t=1}^{T} \left( r_t + \sum_{j=1}^{c} v_{j,t} \right) = \mathcal{R}_T + \sum_{j=1}^{c} \mathcal{V}_{j,T}$$

From Theorem 1 and Theorem 2, we have

$$\sum_{t=1}^{T} \left( r_t + \sum_{j=1}^{c} v_{j,t} \right)$$
$$\leq O(m^2 [\gamma_T T \ln T]^{1/2}) + O(c[T \ln T \gamma_T]^{1/2})$$
$$= O((c + m^2)[T \ln T \gamma_T]^{1/2})$$

Then, by the fact that

$$\min_{\tau \in [T]} \left( r_\tau + \sum_{j=1}^{c} v_{j,\tau} \right) \leq \frac{\sum_{t=1}^{T} \left( r_t + \sum_{j=1}^{c} v_{j,t} \right)}{T}$$

this completes the proof. □

## C PROOFS FOR CONTINUOUS AND COMPACT SEARCH SPACE

### C.1 Assumption

We now consider $\mathcal{X}$ to be continuous and compact. W.l.o.g. $\mathcal{X} := [0, 1]^d$. We keep the assumptions for $\mathcal{GP}$s in §3.2.

### C.2 Generalized CMOBOO

To provide theoretical justification for our algorithm, **COMBOO**, in continuous and compact search spaces, we propose a modified algorithm that is theoretically sound for both finite and infinite search spaces. In this modified algorithm, we slightly adjust the definition of confidence bounds from the main paper while preserving the scalarization method and the approach for declaring infeasibility as outlined in Algorithm 1.

Let $\mathcal{X}$ be a continuous search space. At each iteration $t$, we consider $\bar{\mathcal{X}}_t$, a finite discretization of $\mathcal{X}$, where $\bar{\mathcal{X}}_t$ consists of points evenly distributed across $\mathcal{X}$, with $1/\tau_t$ representing the distance between any two adjacent points in $\bar{\mathcal{X}}_t$. We denote $[\mathbf{x}]_t$ as the closest point in $\bar{\mathcal{X}}_t$ to $\mathbf{x}$.

Using this discretization, we redefine the upper and lower confidence bounds for the objective functions $f_i$ and the constraint functions $g_j$.

**Definition C.1** (Modified confidence bound). For $h \in \{f_i\}_{i \in [m]} \cup \{g_j\}_{j \in [c]}$ define

$$\bar{l}_{h,t}(\mathbf{x}) = \mu_{h,t-1}([\mathbf{x}]_t) - \beta_{h,t}^{\frac{1}{2}} \sigma_{h,t-1}([\mathbf{x}]_t) - \frac{1}{t^2} \quad (33)$$

$$\bar{u}_{h,t}(\mathbf{x}) = \mu_{h,t-1}([\mathbf{x}]_t) + \beta_{h,t}^{\frac{1}{2}} \sigma_{h,t-1}([\mathbf{x}]_t) + \frac{1}{t^2} \quad (34)$$

$\mu_{h,t-1}, \sigma_{h,t-1}$ are defined as in §3.2.

We chose a different confidence parameter for each objective and constraint function $h \in \{f_i\}_{i \in [m]} \cup \{g_j\}_{j \in [c]}$.

$$\beta_{h,t} = 2 \log \left( 2\pi_t (m + c)/\delta \left[ dt^2 b_h \log \left( 2d a_h (m + c)/\delta \right) \right]^d \right)$$

for some constants $a_h, b_h > 0$ and $\pi_t = \frac{\pi^2 t^2}{6}$ $\tau_t = dt^2 B \sqrt{\log(2dA(m + c)/\delta)}$. Detailed definition of $A, B$ are in Paria et al. (2020), B.2.

Lemma 7 serves a similar purpose as Lemma 1 in the discrete case, providing a probabilistic guarantee that the bounds defined above represent the true bounds for our objective and constraint functions.

**Lemma 7** (Lemma 5.7 in Srinivas et al. (2010)). With the assumption for continuous search space and the assumption of the GP model $\forall \delta \in (0, 1), \forall \mathbf{x} \in \mathfrak{F}, \forall h \in \{f_i\}_{i \in [m]} \cup \{g_j\}_{j \in [c]}, \forall t \in [T]$, we have

$$|\mu_{h,t-1}([\mathbf{x}]_t) - h(\mathbf{x})| \leq \beta_{h,t}^{\frac{1}{2}} \sigma_{h,t-1}([\mathbf{x}]_t) + \frac{1}{t^2} \quad (35)$$

with probability $\geq 1 - \delta$.

In the following, we first establish a modified version of Lemma 4, which provides a probabilistic bound on the expected cumulative regret in the continuous case. From this modified lemma, we can derive a bound on the Bayesian regret using a similar approach as in the proof of Theorem 1, ultimately leading to a bound on the cumulative hypervolume (HV) regret in the continuous case when running Algorithm 2. Finally, we will demonstrate three additional results analogous to Theorem 2, Theorem 3, and Theorem 4 when running Algorithm 2.

### C.3 Cumulative Hypervolume Regret Bound

We will show a modified Lemma 4 still holds with confidence bounds replaced by Definition C.1 and Lemma 1 replaced by Lemma 7.

**Algorithm 2** Generalized **COMBOO**

1: $\bar{u}_{g_j,t}$ is defined in Definition C.1
2: **for** $t \in [T]$ **do**
3:     **if** $\max_{\mathbf{x} \in \mathcal{X}}\{\min_{j \in [c]} \bar{u}_{g_j,t}(\mathbf{x})\} < 0$ **then**
4:         **Declare infeasibility.**
5:     **end if**
6:     Sample $\theta_t$ uniformly from $\mathcal{S}_{m-1}^+$.
7:     Find $\mathbf{x}_t \in \arg\max_{\mathbf{x} \in \mathcal{X}} s_{\theta_t}(U_t(\mathbf{x}))$
8:     s.t. $U_t(\mathbf{x}) = (\bar{u}_{f_1,t}(\mathbf{x}) - z_1, ..., \bar{u}_{f_m,t}(\mathbf{x}) - z_m)$
9:     subject to $\bar{u}_{g_j,t}(\mathbf{x}) \geq 0, \forall j \in [c]$.
10:    Evaluate $F, G$ at $\mathbf{x}_t$.
11:    Update GP posterior with the incoming evaluations.
12: **end for**

**Lemma 8** (A modified version of Lemma 4). *In Algorithm 2, suppose $s_\theta(y)$ is L-Lipschitz for all possible $\theta$. With the conditions in Definition C.1 and Lemma 7. For $\delta \in (0,1)$, the expected cumulative regret Definition 13 is bounded with probability at least $1 - \delta$*

$$E[R_C(T)] = O(Lmd^{1/2}[\gamma_T T \ln(T)]^{1/2}) \quad (36)$$

*where $\gamma_T$ is defined in Definition A.1.*

*Remark.* Note that $\gamma_T$ arises because a similar conclusion, used to bound the kernel variance via mutual information gain (MIG), as in Lemma 6, was applied in the proof. However, in Algorithm 2, the input to the variance function $\sigma_{h,t}(\cdot)$ is mapped from a discretized space $\bar{\mathcal{X}}_t$. As a result, $\sum_{t=1}^T \sigma_{h,t-1}([\mathbf{x}]_t)$ should be bounded by $\sqrt{4(T+2)\bar{\gamma}_{h,T}}$, where $\bar{\gamma}_{h,T} = \max_{A_t \subset \bar{\mathcal{X}}_T, |A_t| = t} \frac{1}{2} \log \det(I + \lambda^{-1} K_{h,t})$. The analysis of such MIG with a discretized search space was provided in Lemma 7.5 of Srinivas et al. (2010), demonstrating that $\gamma_{h,T} \geq \bar{\gamma}_{h,T}$. For consistency, we use the MIG over the entire search space, $\gamma_{h,T}$, rather than $\bar{\gamma}_{h,T}$, in the subsequent results.

*Proof of Lemma 8.* In this proof, we employ a different strategy by splitting the expected cumulative regret into three components, rather than the two in proof of Lemma 4, because we bound $R_C(T)$ by an extra new error term from the discretization of search space. We will provide separate bounds for each of these components.

Define $\mathfrak{F}_t = \{\mathbf{x} \mid \bar{u}_{g_j,t}(\mathbf{x}) \geq 0, \forall j \in [c]\}$. Then, take $\mathbf{x}_t^* = \arg\max_{\mathbf{x} \in \mathfrak{F}_t} s_{\theta_t}(F(\mathbf{x}))$ and define $\mathbf{x}_t = \arg\max_{\mathbf{x} \in \mathfrak{F}_t} s_{\theta_t}(U_t(\mathbf{x}))$. It follows that

$$E[R_C(T)] = E\left[\sum_{t=1}^T \left(\max_{\mathbf{x} \in \mathfrak{F}} s_{\theta_t}(F(\mathbf{x})) - s_{\theta_t}(F(\mathbf{x}_t))\right)\right]$$

$$\leq E\underbrace{\left[\sum_{t=1}^T s_{\theta_t}(U_t(\mathbf{x}_t)) - s_{\theta_t}(F(\mathbf{x}_t))\right]}_{B_1} +$$

$$E\underbrace{\left[\sum_{t=1}^T s_{\theta_t}(F([\mathbf{x}_t^*]_t)) - s_{\theta_t}(U_t([\mathbf{x}_t^*]_t))\right]}_{B_2} +$$

$$E\underbrace{\left[\sum_{t=1}^T s_{\theta_t}(F(\mathbf{x}_t^*)) - s_{\theta_t}(F([\mathbf{x}_t^*]_t))\right]}_{B_3} \quad (37)$$

when

$$s_{\theta_t}(U_t(\mathbf{x}_t)) \geq s_{\theta_t}(U_t(\mathbf{x}_t^*))$$
$$s_{\theta_t}(U_t(\mathbf{x}_t)) \geq s_{\theta_t}(U_t([\mathbf{x}_t^*]_t))$$

and

$$\max_{\mathbf{x} \in \mathfrak{F}_t} s_{\theta_t}(F(\mathbf{x})) \geq \max_{\mathbf{x} \in \mathfrak{F}} s_{\theta_t}(F(\mathbf{x}))$$

The first two conditions hold due to the choice of $\mathbf{x}_t$. From Lemma 7, we know $\bar{u}_{g_j,t}(\mathbf{x}) \geq g_j(\mathbf{x}), \forall j \in [c], t \in [T], \mathbf{x} \in \mathcal{X}$ with probability at least $1 - \delta$, then $\mathfrak{F} \subset \mathfrak{F}_t$ with probability at least $1 - \delta$, which means the third condition holds with the same probability.

We can follow the Proof of Lemma 1, 2, 3 in Paria et al. (2020) to show three terms in (37) are bounded. Take $\beta_T$ as an upper bound of $\beta_{f_i,T}, \forall i \in [m]$. For $B_1, B_2$, we can apply the same bound as in (16) and (17).

$$B_1 \leq \left[L(m\beta_T T)^{1/2} \left(\sum_{i=1}^m \frac{\gamma_{f_i,T}}{\ln(1+\sigma^{-2})}\right)^{1/2}\right]$$
$$+ Lm\frac{\delta}{2(m+c)\tau_t^d}$$

$$B_2 \leq Lm \sum_{t=1}^T \sum_{\mathbf{x} \in \bar{\mathcal{X}}_t} \exp(-\frac{\beta_{f_i,t}}{2}) \leq Lm\frac{\delta}{2(m+c)}$$

By result in Ghosal and Roy (2006), if kernel $k$ is stationary and $4^{th}$-differentiable, $h \sim \mathcal{GP}$. $\exists a_h, b_h > 0$ s.t. $\forall k \in \{1, ..., d\}$

$$\mathbb{P}\left(\sup_{\mathbf{x}} \left|\frac{dh}{d\mathbf{x}_k}\right| > L\right) \leq a_h \exp\left((-L/b_h)^2\right) \quad (38)$$

Here we define $A, B$ in Lemma 7:

$$A = \sup_{h \in \{f_i\}_{i \in [m]} \cup \{g_j\}_{j \in [c]}} a_h$$

$$B = \sup_{h \in \{f_i\}_{i \in [m]} \cup \{g_j\}_{j \in [c]}} b_h$$

From equation 16 of Paria et al. (2020)

$$B_3 \leq \sum_{t=1}^{T} Lm \frac{dAB\sqrt{\pi}}{2\tau_t} \quad (39)$$

Since $\tau_t \sim O(dt^2)$, $B_3 \leq O(Lm)$.

Finally, we conclude that for some global constants $C_1, C_2 > 0$, expected cumulative regret is bounded in the following.

$$E[R_C(T)] \leq C_1 Lm \quad (40)$$

$$+ C_2 L \left( mT(d \ln T + d \ln d) \sum_{i=1}^{m} \frac{\gamma_{f_i,T}}{\ln(1 + \sigma^{-2})} \right)^{1/2} \quad (41)$$

with probability at least $1 - \delta$. The final conclusion follows from (41). $\square$

### C.4 Cumulative Constraint Violation Bound

**Lemma 9.** *With the conditions in Lemma 7,*

$$v_{g_j,t} \leq 2\beta_{g_j,t} \sigma_{j,t-1}([\mathbf{x}_t]_t) + \frac{2}{t^2}, \; \forall j \in [c], \forall t \in [T] \quad (42)$$

*we have with probability at least $1 - \delta$.*

The proof Lemma 9 is analogous to the proof of Lemma 5. From Lemma 9, by taking the summation over $t$, we can bound cumulative constraint violation by $4\sqrt{\beta_{g_j,T}(T+2)\gamma_{g_j,T}} + \frac{\pi^2}{3} \leq O(\sqrt{T \ln T \gamma_{g_j,T}})$

### C.5 Declaration of Infeasibility

The conclusion of Theorem 3 is consistent in the continuous setting. The proof is the same as the discrete case. Finally, given the problem is feasible, we can conclude $\arg\max_{\mathbf{x} \in \mathcal{X}} \min_{j \in [c]} \bar{u}_{g_j,t}(\mathbf{x}) \geq 0$ holds with probability at least $1 - \delta$.

For Theorem 4, similarly, we define $\epsilon$ as $|\max_{\mathbf{x} \in \mathcal{X}} \min_{j \in [c]} g_j(\mathbf{x})|$. Replace $u_{g_j,t}, l_{g_j,t}$ by $\bar{u}_{g_j,t}, \bar{l}_{g_j,t}$. Following the same strategy, we will have

$$\epsilon \leq \frac{2\beta_T^{\frac{1}{2}}\sqrt{4(T+2)\gamma_T}}{T} + \frac{\pi^2}{3T} \leq O(\frac{\beta_T^{1/2}\sqrt{\gamma_T}}{\sqrt{T}}) \quad (43)$$

with high probability.

## D EXPERIMENTAL DETAILS

### D.1 Implementation

We follow the tutorial for **qNEHVI** and **qParEGO** found at https://botorch.org/tutorials/constrained_multi_objective_bo to implement those benchmarks.

The implementation of **MESMOC** we tested can be found at https://github.com/belakaria/MESMOC.

The implementation and datasets for **COMBOO** can be found at https://github.com/dancewithDianTong/COMBOO.

### D.2 Test Functions

**Toy Function**

The objective is defined as:

$$F(\mathbf{x}_1, \mathbf{x}_2) = \left( -\frac{1}{\mathbf{x}_1} - \mathbf{x}_2, -\mathbf{x}_1 - \mathbf{x}_2^2 \right)$$

$$\text{s.t.} \quad \mathbf{x}_1, \mathbf{x}_2 \in [1, 1.5]$$

$$-\frac{1}{\mathbf{x}_1} - \mathbf{x}_2 \geq -1.9, -\mathbf{x}_1 - \mathbf{x}_2^2 \geq -2.25$$

Let $\beta_{g_j,T} = 0.4 \log(4 \cdot (1 + t))$. We used the Matérn kernel for the GP model with 0.05 standard deviation noise. We took ten random initial candidates in each trial.

$$k_{\text{Matérn}}(\mathbf{x_1}, \mathbf{x_2}) = \frac{2^{1-\nu}}{\Gamma(\nu)} \left( \sqrt{2\nu}d \right)^{\nu} K_{\nu}\left( \sqrt{2\nu}d \right)$$

- $d = (\mathbf{x_1} - \mathbf{x_2})^\top \Theta^{-2} (\mathbf{x_1} - \mathbf{x_2})$ is the distance between $\mathbf{x_1}$ and $\mathbf{x_2}$, scaled by the lengthscale parameter $\Theta$.

- $\nu$ is a smoothness parameter that takes values $\frac{1}{2}$, $\frac{3}{2}$, or $\frac{5}{2}$. Smaller values correspond to less smoothness.

- $K_\nu$ is a modified Bessel function.

**Branin-Currin**

A 2-D objective consisting of a Branin function and a Currin function.

$$f_1 = 15\mathbf{x}_2 - \left(5.1 \cdot \frac{(15\mathbf{x}_1 - 5)^2}{4\pi^2} + \frac{5(15\mathbf{x}_1 - 5)}{\pi} - 5\right)^2$$
$$+ \left(10 - \frac{10}{8\pi}\right)\cos(15\mathbf{x}_1 - 5)$$
$$f_2 = \left(1 - \exp\left(-\frac{1}{2\mathbf{x}_2}\right)\right)$$
$$\cdot \left(\frac{2300\mathbf{x}_1^3 + 1900\mathbf{x}_1^2 + 2092\mathbf{x}_1 + 60}{100\mathbf{x}_1^3 + 500\mathbf{x}_1^2 + 4\mathbf{x}_1 + 20}\right)$$
s.t. $\mathbf{x}_1, \mathbf{x}_2 \in [0, 1]$
$f_1 \geq -20, f_2 \geq -6$

We let $\beta_{g_j,T} = 0.4 \log(4 \cdot (1+t))$. We use the Matérn kernel with 0.01 standard deviation and ten random initial candidates in each trial.

**C2-DTLZ2**

$$f_i(\mathbf{x}) = (1 + g(\mathbf{x}_m))\cos\left(\frac{\pi}{2}x_i\right), \forall i \in [m]$$

where $g(\mathbf{x}) = \sum_{x_i \in \mathbf{x}_m}(x_i - 0.5)^2, \mathbf{x} \in [0,1]^d$, and $\mathbf{x}_m$ represents the last $d - m + 1$ elements of $\mathbf{x}$. The constraint of this problem is defined as: $c(\mathbf{x}) = -\min\left[\min_{i=1}^m\left((f_i(\mathbf{x}) - 1)^2 + \sum_{j=1, j\neq i}^m (f_j^2 - r^2)\right), \left(\sum_{i=1}^m\left((f_i(\mathbf{x}) - \frac{1}{\sqrt{m}})^2 - r^2\right)\right)\right] \geq 0$ where $\mathbf{x} \in [0,1]^d$ and $r = 0.2$. In our case, $m = 2, d = 2$. We used the same setting for the GP model and $\beta_{g_j,T}$ as the **Toy Function**. For this objective, we imposed a feasibility condition on the initial sampled points across all experiments. Given the complexity of the constraint for this objective, **COMBOO** 's auxiliary problem is likely to declare infeasibility when dealing with random initial samples. However, since the objective is known to be feasible, we argue that there is no need to discard the estimations aimed at detecting infeasibility. Instead, these estimations should be leveraged to explore the feasible region effectively.

**Disc Brake Design Problem**

It is the same problem with RE-3-4-3 in the real-world constraint multi-objective problem set Tanabe and Ishibuchi (2020). We use the same setting for the GP model and $\beta_{g_j,T}$ as the **Toy Function**.

**Caco-2++**

The original version **Caco-2+**(d = 2133, m = 3) was proposed in Park et al. (2024), whose objective contains *permeability*, an experimentally tested value and two extra objectives, CrippenClogP, TPSA. The search space contains 906 drug molecules. We modified this dataset to come up with **Caco-2++**(d = 2175, m = 3, c = 3). For the domain, we augmented a new feature, mqn feature, to the domain of **Caco-2+**. We changed the objectives to *permeability*, TPSA, and drug-likeliness score(QED). The search space contains 909 molecules. We constrain the objectives so that QED $\geq 0.5$, TPSA $\geq 80$, *permeability* $\geq -5$.

We took $\beta_{g_j,T} = c \log(2(1+t))$, $c = 0.1$ or $0.05$. We used a Tanimoto kernel specialized for molecule representation. We take a 0.01 standard deviation observation noise and 64 initial candidates in each trial.

$$k_{\text{Tanimoto}}(\mathbf{x}_1, \mathbf{x}_2) = a \cdot \frac{\mathbf{x}_1 \cdot \mathbf{x}_2}{\|\mathbf{x}_1\|^2 + \|\mathbf{x}_2\|^2 - \mathbf{x}_1 \cdot \mathbf{x}_2}$$

where $a$ is the signal amplitute parameter, and was set to be 1.

**ESOL+**

The original **ESOL** (Delaney (2004)) dataset contains 1,144 organic molecules and an experimentally measured metric called log*(Solubility)*. We added three additional objectives—LogP, TPSA, and QED—to the original objective. For the domain, we use the same molecule representation, *fragprint*, as was used by **Caco-2+** (d = 2133). We constrain the objectives so that LogP $\geq 2.5$, QED $\geq 0.5$, TPSA $\geq 55$, and log*(Solubility)* $\geq -4$.

We took $\beta_{g_j,T} = c \log(2(1+t))$, $c = 0.1$ or $0.05$. We used a Tanimoto kernel and a 0.005 standard deviation observation noise and 64 initial candidates in each trial.

**Penicillin Function**

The objective was proposed in Liang and Lai (2021). We add a Gaussian noise with a standard deviation of 0.05 to the observations. We define the constraint to make penicillin production $\geq 10$, $CO_2$ production $\leq 60$ and reaction time $\leq 350$.

We took $\beta_{j,t} = c \log(2(t+1))$, where $c = 0.1$ or $c = 0.05$. We used the RBF kernel

$$k_{\text{RBF}}(\mathbf{x}_1, \mathbf{x}_2) = a \cdot \exp\left\{\frac{\|\mathbf{x}_1 - \mathbf{x}_2\|^2}{b}\right\}$$

We fit the GP models' parameters in each step in all experiments.

*Remark.* Due to the highly skewed distribution of the objective, we observed significant overfitting of the GP model, even with large training data. To address this, we applied the Voxel Grid Sampling trick to the initial samples of the GP model for each benchmark. Specifically, we filtered 64 randomly sampled candidates to get around 20 uniformly distributed samples within the initial candidate range for each trial. This adjustment makes the performance of all benchmarks differ from that of a random search.

### D.3 Figures

We present the additional results here.

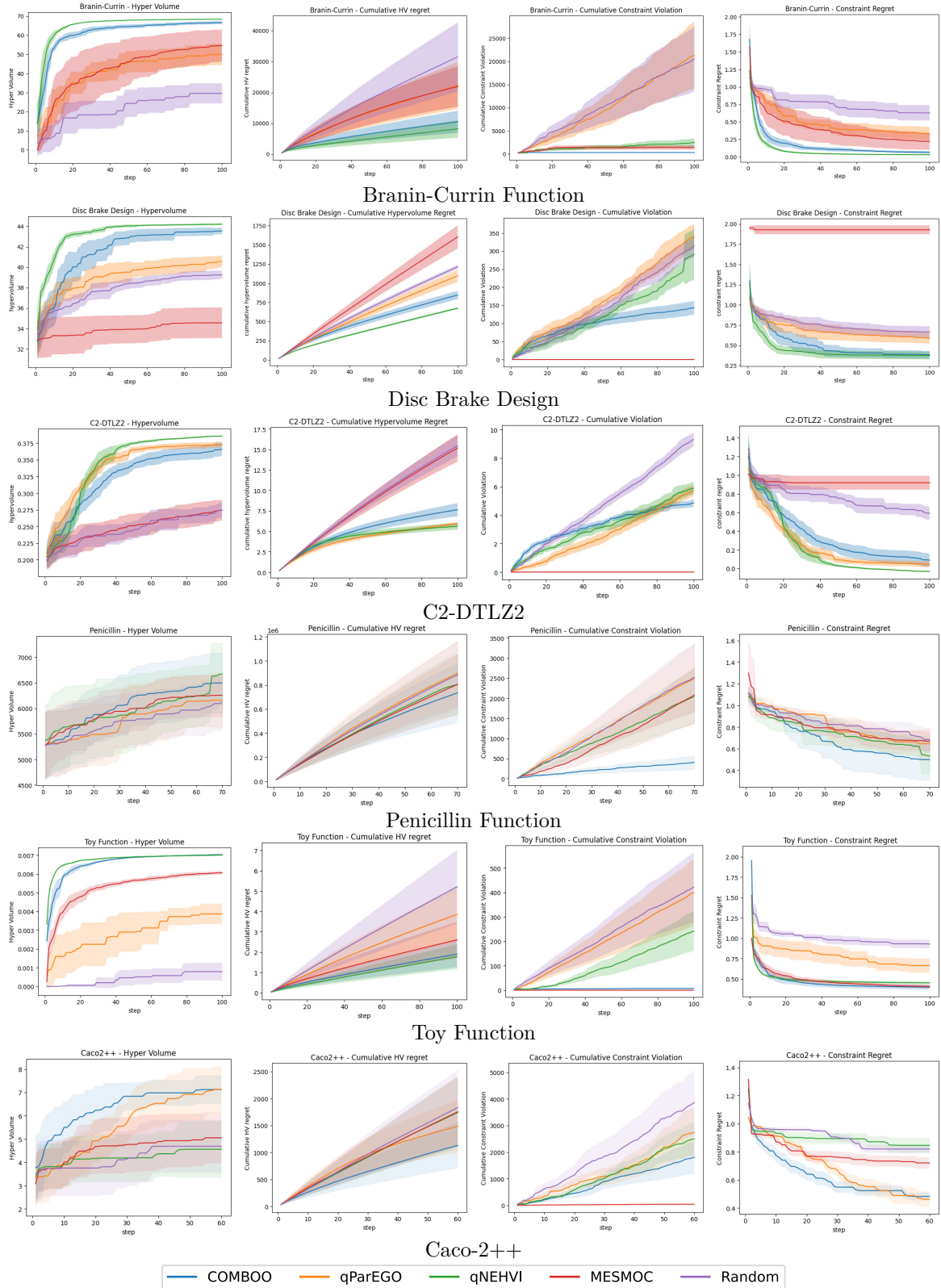

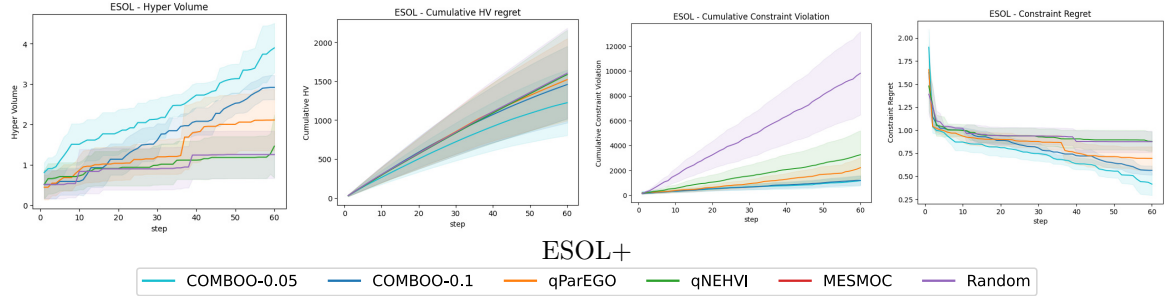

ESOL+

Figure 6: **COMBOO** performance on other objectives. From left to right: Hypervolume, Cumulative Hypervolume Regret, Cumulative Constraint Violation, Constraint Regret. Curves are shaded by area between $\pm$ 1.96 standard error. 0.05 and 0.1 are the coefficients of the confidence parameter $\beta_t$.

As shown in Figure 6, **COMBOO** consistently matches the performance of the baselines, particularly qParEGO and qNEHVI, in terms of hypervolume. Additionally, it significantly reduces violations. MESMOC, on the other hand, tends to avoid violations through conservative feasibility estimation, as depicted in Figure 1, but falls behind in hypervolume improvement. Overall, **COMBOO** represents a principled trade-off between feasibility exploration and multi-objective optimization, demonstrating consistent and competitive performance across various tasks.